# A Compact Self-organizing Cellular Automata-based Genetic Algorithm

Vasileios Barmpoutis                                                                                    vb28@buffalo.edu
Department of Economics, University at Buffalo, State University of New York, Buffalo, NY 14260 USA

Gary F. Dargush                                                                                    gdargush@eng.buffalo.edu
Department of Mechanical and Aerospace Engineering, University at Buffalo, State University of New York, Buffalo, NY 14260 USA

**Abstract**

A Genetic Algorithm (GA) is proposed in which each member of the population can change schemata only with its neighbors according to a rule. The rule methodology and the neighborhood structure employ elements from the Cellular Automata (CA) strategies. Each member of the GA population is assigned to a cell and crossover takes place only between adjacent cells, according to the predefined rule. Although combinations of CA and GA approaches have appeared previously, here we rely on the inherent self-organizing features of CA, rather than on parallelism. This conceptual shift directs us toward the evolution of compact populations containing only a handful of members. We find that the resulting algorithm can search the design space more efficiently than traditional GA strategies due to its ability to exploit mutations within this compact self-organizing population. Consequently, premature convergence is avoided and the final results often are more accurate. In order to reinforce the superior mutation capability, a re-initialization strategy also is implemented. Ten test functions and two benchmark structural engineering truss design problems are examined in order to demonstrate the performance of the method.

## 1    Introduction

Nature inspired methods have attracted great interest in recent years. The most prominent representatives are Genetic Algorithms, Evolutionary Programming, Evolutionary Strategies and Genetic Programming. These new approaches have certain advantages in comparison with the more traditional methods of optimization. In particular, the performance of traditional methods deteriorates significantly when the problem becomes complicated. Additionally, traditional methods usually require gradient information in order to move towards the optimal solution.

In this work, a new Genetic Algorithm (GA) is proposed. GAs are stochastic methods of optimization based on the Darwinian principle of natural selection. These methods can handle discontinuities and non-convex regions, and, in general, do not require gradient information. Consequently, GAs are very general methods with a broad range of applicability. However, GAs have certain disadvantages as well, including an inability to take constraints directly into consideration, premature convergence, large computational time, and lack of precision in the final solution.

The original development of GAs was by Holland (1975). The basic approach also is described in the monographs by Goldberg (1989) and Mitchell (1996), which include many applications. In recent years, a number of researchers have proposed improvements to the standard GAs. These improved versions focus on an enhanced quality of local search (e.g., Ishibuchi and Murata, 1998; Guo and Yu, 2003; Cui et al., 2003), methodologies to enhance the overall performance of GAs and avoid premature convergence (e.g., Krishnakumar, 1989; Koumousis and Katsaras, 2006) and improved strategies to enforce the constraints of the problem (e.g., Venkatraman and Yen, 2005). However, the basic limitations remain and further improvements are necessary to enhance performance.

Cellular Automata (CA) represent a relatively new approach to problems of significant difficulty in the analysis of natural phenomena. Traditionally, these problems are formulated using mathematical equations (usually differential equations). However, for systems with organized complexity, the analytical solution of these equations becomes intractable and alternative approaches are of interest (Weaver, 1948). In general, CA employ simple rules to investigate complicated systems whose study may be difficult with traditional means. The novelty of CA is the use of local rules, which allow the system to develop a certain behavior without the explicit formulation of a (global) mathematical equation to govern its behavior. CA are very flexible and thus powerful tools. However, the heuristic development of local rules often involves great difficulty and an inappropriate rule can drive the system toward completely false behavior.

CA were first used to study heart cells by Wiener and Rosenbluth (1946), while the initial theoretical studies of CA were conducted by Ulam (1952) and von Neumann (1966). Since that time, many researchers have contributed significantly to the theory and application of CA. For example, Wolfram (1994) investigated CA with the help of statistical physics and reached several conclusions that explain CA behavior. One recent application where CA can be used instead of traditional differential equations can be found in Kawamura et al. (2006). The authors employ a finite difference approach to express the wave equation in an iterative and localized form and then use CA to represent this localized structure.

Early research on combining ideas from Cellular Automata with Genetic Algorithms includes the work by Manderick and Spiessens (1989), Gorges-Schleuter (1989), Hillis (1990), Collins and Jefferson (1991), Davidor (1991), Mühlenbein (1991), Whitley (1993) and Tomassini (1993). These and related approaches all can be viewed within the general framework of Parallel Genetic Algorithms (PGA), many of which are surveyed in the review paper by Alba and Tomassini (2002). Recent focus has been toward understanding the performance of PGAs and Cellular Genetic Algorithms (CGAs). For example, Sarma and DeJong (1996) studied the growth curves as a function of the ratio of the radius of the neighborhood to the radius of the grid and a logistic curve is used to approximate the growth curve. In Rudolph (2000), takeover time for arbitrary structures, arrays and rings are calculated for Cellular Evolutionary Algorithms (CEAs). More recently, the growth curves and takeover times for distributed genetic algorithms are studied in Alba and Luque (2004) and three theoretical models are tested for fitting the growth curves under different migration frequencies and rates. Meanwhile, in an interesting study, Alba and Troya (2000) show that the 2-D grid can be used for both exploration and exploitation by altering the grid dimensions and the neighborhood. Sipper et al. (1998) investigated the evolution of non-uniform CA with regard to asynchrony and scalability. Relative performance of synchronous and asynchronous updating approaches are discussed in Schönfisch and de Roos (1999) within the context of CA. In Alba et al. (2002), three asynchronous CEAs are used and the asynchronous policies are tested for three different problems. Giancobini et al. (2005) provides a survey on the selection intensity of different asynchronous policies with respect to synchronous updates and panmictic methodology and creates theoretical models for the selection pressure in regular lattices under synchronous and asynchronous updates. Finally, we mention the work by Suzudo (2004), where the author used a GA to investigate spatial patterns of CA. However, it should be noted that all of these efforts are focused toward development of massively parallel approaches.

The remainder of this paper is organized as follows. The overall conceptual basis for the proposed algorithm is first presented in Section 2 to provide motivation and to clarify its distinctive features. Then, in Section 3, complete implementation details of this compact self-organizing CA-GA are provided. In order to examine the performance of the proposed algorithm, the optimization of ten test functions is considered in Section 4. Afterwards, in Section 5, the new approach is applied to two problems of structural optimization. Section 6 contains conclusions and some final thoughts on the potential applicability of the method.

## 2  Conceptual Basis for the Proposed Algorithm

Genetic algorithms employ a population of solutions as an initial seed and then, with the use of selection, crossover and mutation, they evolve to produce improving solutions. GAs owe their power mainly to selection and crossover operators, while the mutation operator has only secondary significance, i.e,. to randomly search for better solutions in the domain. The low significance of mutation is evident from the low mutation rates used in most GAs.

Selection and crossover have the tendency to organize the initial solutions (Eiben et al., 1991; Prügel-Bennett and Shapiro, 1994; Leung et al., 1997; Suzudo, 2004), so after some generations the solutions are improved in average and the diversity of the population is lost. Consequently, when crossover is the main operator, the initial pool of solutions should be large enough to ensure a large diversity of the initial solutions and a complete coverage of the search domain. On the other hand, mutation tends to have the opposite effect. The diversity tends to increase with mutation and the domain can be searched in a more complete way, but the average of the solutions is not improved. One possible reason that mutation has such an effect is that not only does it provide randomness per se, but also that this randomness is free to spread itself through the global crossover scheme used in GAs.

The distractive effects of high mutation rates can be shown through a simple example. Consider the minimization of the following function:

$$f = \sum_{i=1}^{N} x_i^2 \qquad (1)$$

for $-50 \leq x_i \leq 50$. In Figure 1, the average minimum from 30 independent runs is shown as a function of the mutation rate per bit for $N=15$ with five bits employed to encode each variable. For the simulations, the PIKAIA algorithm (Charbonneau and B. Knapp, 2007) was used by evaluating 100 individuals over 100 generations (i.e., 10000 function evaluations) with a crossover probability of 0.85. It is obvious that an increase in mutation rate does not allow crossover to play its role in increasing the average of the solutions. However, in many problems, a lack of mutation increases the chance of premature convergence of the GA (Leung et al., 1997).

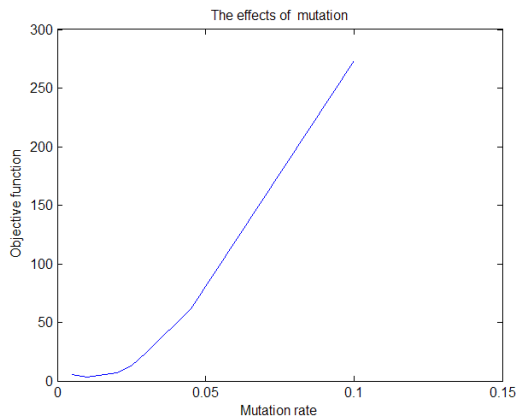

Figure 1: The effects of increasing mutation rate for minimizing (1) averaged over 30 independent runs.

Cellular automata use localized structures to solve problems in an evolutionary way. CA often demonstrate also significant ability toward self-organization that comes mostly from the localized structure on which they operate. By organization, one means that after some time in the evolutionary process, the system exhibits more or less stable localized structures (Wolfram, 2002). This behavior can be found no matter the initial conditions of the automaton. Of course, not all CA organize themselves. For example, Wolfram (2002) identifies four classes of CA with increasing levels of complexity. The first two classes encompass relatively simple behavior in which the system evolves to either a uniform (class 1) or non-uniform (class 2) steady state. On the other hand, in class 3 and especially class 4, a change in initial conditions can produce changes in the evolving patterns. However, we still can see the same localized structures, even though these structures may have different locations and scales. Another important feature of class 4 systems is that these localized structures can move in the automaton, while the automaton itself is not disturbed in the procedure.

In order to illustrate these ideas, consider next the simple rule displayed in Figure 2. The evolution of the rule is shown in Figure 3 for two different random initial conditions. It is obvious that the rule creates distinct triangular-like patterns. For different initial conditions, we see that the patterns always persist, but in different locations. Finally, we can introduce random changes in the middle of the evolution. In Figure 4, an average of 10 cells change state randomly in every step. The position of the modified cells is random, as is the initial state of the system. Yet, we see again that a similar triangular pattern emerges.

```
111   110   100   101   011   010   001   000
 1     0     0     1     0     1     0     1
```

Figure 2: A simple CA rule.

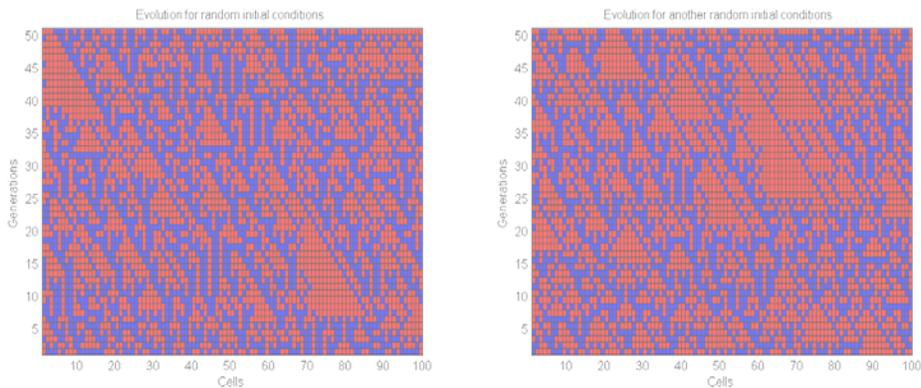

Figure 3: CA rule evolution for 50 generations with different initial seeds.

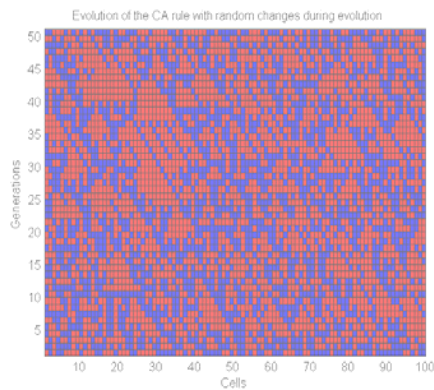

Figure 4: CA rule evolution for 50 generations with random changes during the evolution.

From the above results, we can conclude that a GA based on a CA methodology could be more stable and sustain higher rates of mutation. The idea is that the distractive effects of mutation can be offset by the additional self-organization and stability that the CA can introduce in the system. Another effect of the higher mutation rates is that the initial pool of solutions does not need to be as large as in traditional GAs. Thus, it should be emphasized that, unlike existing cellular genetic algorithms, the proposed algorithm does not focus on the inherent CA parallelism, but rather on their self-organizing characteristics.

## 3  Details of the Proposed Algorithm

The proposed algorithm creates a CA framework for GAs. Every individual of the population is assigned to a cell. Therefore, the number of CA lattice cells is equal to the number of individuals in the population. Additionally, an iteration of the CA now corresponds to a generation of the GA. However, the proposed approach abandons the global statistics that control the evolution of the population in a standard GA. Instead, local rules direct the evolution on the population lattice.

In the present implementation, a 1-D lattice is utilized, but a 2-D or N-D lattice also could be employed. For the simple 1-D case, every internal cell communicates with the two adjacent cells only. At each generation, every interior cell is compared with the two other cells that form its neighborhood. Five distinct cases exist: 1) the cell has a higher fitness than either neighboring cell, 2) the cell has a better fitness than the left cell only, 3) the cell has better fitness than the right cell only, 4) the cell has the worst fitness among its neighbors and 5) all the cells have the same fitness. During the generation, the crossover operation depends on these cases: For case 1, the cell remains intact and it survives unmodified through to the next generation (Figure 5a). For cases 2 or 3, crossover takes place between the individual of the central cell and the individual of the right cell (Figure 5b) or left cell (Figure 5c), respectively. For case 4, the crossover on this cell takes place between the individual of the left and right cell and nothing survives from the individual that possessed the central cell at the beginning of the current generation (Figure 5d). Finally, for case 5, nothing happens and the central cell survives to the next generation. For boundary cells on the two ends of the lattice, communication is limited only to the single adjacent cell and the boundary cell uses only this single adjacent cell for fitness comparison and exchange of substrings.

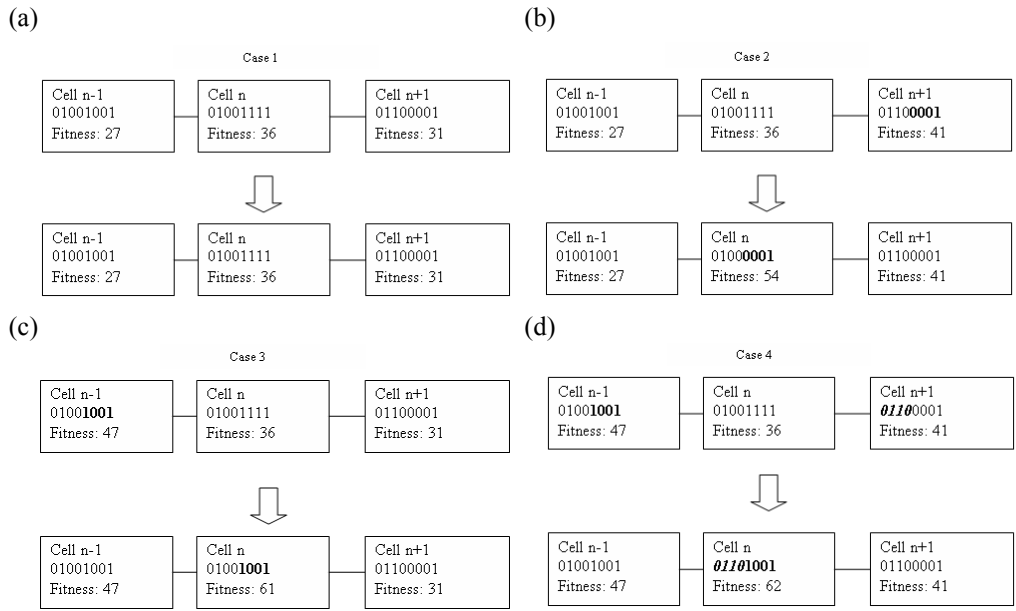

Figure 5: Evolution rules for the proposed algorithm.

The algorithm reads the cells from left to right. The process starts with the comparison of the first individual with its right neighbor. When this operation is finished, the next cell (the second individual from the left) is compared with its right (the third cell of the lattice) and left (the first cell) neighbor. During this process, the old string and fitness are used for the left cell, not the updated values. This holds for all cases; the selection and crossover operations always utilize the state of the left cell at the beginning of the present generation, rather than the updated state. Thus, the proposed CA framework is synchronous, which proves to be beneficial in conjunction with high mutation rates.

From the description above, it is easily understood that the global statistics of all individuals are abolished. Meanwhile, there are three kinds of crossover that can be used: one point crossover, two point and variable-to-variable crossover. Furthermore, a combination of crossovers also can be used at different stages of the optimization process.

In the proposed methodology, mutation has an increased importance compared to other GAs and, consequently, three different kinds of mutation are introduced. The first kind is the regular mutation in which, after crossover, a bit of information is altered. The second kind of mutation has to do with imposing mutation on the best individual of a generation. The third kind is a hyper-mutation that changes whole substrings of individuals by parts of individuals from past generations. Each mutation type has a different probability of occurrence. The first two kinds of mutation each have two different versions: the first version imposes random changes to the strings of the individual (ordinary mutation). When the ordinary version of the regular mutation occurs, one or more bits of the substrings of every variable are changed. So, if the strings of the population consist of three substrings which encode the information of three variables, then three mutations will take place (i.e., at least one for every substring). However, the substrings do not need to belong to the same individual. For example, the first mutation may happen to the first substring of the fifth individual, the second mutation in the second substring of the $14^{th}$ individual and the third mutation to the third substring of the $21^{st}$ individual. The number of times the procedure repeats can be predefined or it can be a random number (i.e., different each time). The ordinary version of the mutation of the best individual changes a number of bits within the string. The number of bits is a random number with an upper bound equal to the number of substrings of the individual. However, the number of bits to change can be set to any number and the mutations do not have to occur in different substrings. The second version of mutation adds a random value to a substring of an individual (Gaussian mutation). The regular version of this Gaussian mutation can be repeated more than once, while in the case of mutation in the best individual, it happens only one time.

Another feature of the proposed methodology is reinitialization. Reinitialization was introduced by microGAs (Krishnakumar, 1989) and has been shown to give improved results. During the initialization discussed here, the algorithm is not restarted randomly; all the cells lose their previous states and the best-so-far individual of all previous generations is introduced to all cells. Therefore, all cells for the next generation have the same individual; that is, the best individual of all the previous generations. The interval of reinitialization can be constant or not. Usually, reinitialization does not happen from the very beginning of a run, but only after the algorithm has executed a number of generations.

## 4   Benchmark Examples

Next, to examine the efficacy of the proposed algorithm, a number of example optimization problems are considered. In all cases, problem parameters are encoded as real-valued, rather than the binary encoding discussed above. As a result, a few of the genetic operators require further discussion.

In the case of Gaussian mutation for the real coded GA used here, a random integer number $\iota$ is added to a random digit of the string. Thus,

$$x_i = x_i + \iota \quad (2)$$

where $\iota$ is a normally distributed random integer between 0 and 9. However, the ordinary mutation is still used. Ordinary means that the mutation strategy imitates the one employed with binary coding, where there is random change of a bit. In the case of the present real coded GA, this means that the chosen digit with a value between 0 and 9 changes its value to another integer between 0 and 9. So, if a digit has the value 4, then after the regular mutation it gets another value, e.g. 6. Thus, the value of the substring is altered because the random mutation that has changed one of its digits.

In the remainder of this section, the ten test functions, defined in Table 1, will be considered. The approach followed for the test suite is similar to that utilized in Koumousis and Katsaras (2006). The first two test functions will be examined in detail in order to capture the main properties of the proposed algorithm. The remaining eight functions, along with the best performing cases of the first two functions, will be tested against PIKAIA (Charbonneau and Knapp, 2007), an ordinary GA.

*1$^{st}$ test function*

The impact of population size, rate and type of mutation, frequency of reinitialization and the starting point of reinitialization will be examined for maximization of the first test function defined in Table 1. In order to capture the impact of each factor separately, different values of each will be tried, while the other factors remain constant. Variable to variable crossover was used along with Gaussian mutation. When mutation did occur, only a single mutation was involved. The initial configuration includes the following: a population of five cells over 2000 generations (i.e., 10000 function evaluations), Gaussian mutation of the best individual every two generations, one Gaussian mutation every generation, and reinitialization every three generations beginning after generation 500. This configuration is tested against cases employing populations of 10 and 20 cells. All of the other parameters remain constant, except for the starting point of reinitialization, which occurs after 250 generations in the case with 10 cells and after 125 generations in the case of 20 cells (i.e., in all three cases after 25% of the total function evaluations). The results of these three initial cases, in terms of mean values of the best individual over 60 runs, are shown in Table 2. The numbers in parenthesis represent the standard deviations.

The case with a population of five cells has performed better than the other two cases, while simulations with 10 cells also have performed well. The 20-cell case did not perform nearly as well and, in some cases, it has converged to local optima. The reason for that can be found in the fact that the mutations for the 20-cell case were less than the two other cases; since we have one mutation per generation, and the 20-cell case runs only for 500 generations, in comparison to the 2000 generations for the 5-cell case. This is a first indication that mutation has an important role in the proposed algorithm.

Next, we consider the cases where 1) no reinitialization is performed, 2) no mutation of the best individual is used and 3) no mutation is employed. The results provided in Table 3 are the mean values obtained over 60 runs with the corresponding standard deviations provided in parentheses. From Table 3 we see that omission of any of the three factors has a negative impact on performance, except perhaps for reinitialization in the 20-cell simulations. Notice that the performance of the algorithm degrades substantially, especially when mutation is excluded. This is a second indication that mutation is important for the algorithm.

Table 1: Test function

| | Formula | $N_i$ | $S_i$ | Resolution |
|---|---|---|---|---|
| 1st function (max) | $f = \prod_{i=1}^{N} \left[ (\sin(5.1 x_i + 0.5))^{30} \exp\left( -4 \log(2) \frac{(x_i - 0.0667)^2}{0.64} \right) \right]$ | 5 | $0 \leq x_i \leq 1$ | 0.0001 |
| 2nd function (min) | $f = \sum_{i=1}^{N-1} \left[ 100(x_{i+1} - x_i^2) + (1 - x_i)^2 \right]$ | 3 | $-5 \leq x_i \leq 5$ | 0.00001 |
| 3rd function (min) | $f = 20 + e - 20 \exp\left[ -0.2 \sqrt{\frac{1}{N} \sum_{i=1}^{N} x_i^2} \right] - \exp\left[ \frac{1}{N} \sum_{i=1}^{N} \cos(2\pi x_i) \right]$ | 15 | $-100 \leq x_i \leq 100$ | 0.0002 |
| 4th function (min) | $f = -\sum_{i=1}^{N} x_i \sin\left(\sqrt{|x_i|}\right)$ | 15 | $-500 \leq x_i \leq 500$ | 1.0 |
| 5th function (min) | $f = \frac{1}{4000} \sum_{i=1}^{N} x_i^2 - \prod_{i=1}^{N} \cos\left(\frac{x_i}{\sqrt{i}}\right) + 1$ | 15 | $-500 \leq x_i \leq 500$ | 0.1 |
| 6th function (min) | $f = \sum_{i=1}^{N} \left( x_i^2 - 10 \cos(2\pi x_i) + 10 \right)$ | 15 | $-5 \leq x_i \leq 5$ | 0.001 |
| 7th function (min) | $f = \sum_{i=1}^{N} x_i^2$ | 15 | $-50 \leq x_i \leq 50$ | 0.001 |
| 8th function (max) | $f = \sum_{i=1}^{N} \left( \prod_{i=1}^{j} x_i \right) \left( \cos\left( \prod_{i=1}^{j} x_i \right) \right)$ | 4 | $0 \leq x_i \leq 100$ | 0.0001 |
| 9th function (max) | $f = x_i \sin(10 \pi x_i)$ | 15 | $0 \leq x_i \leq 10$ | 0.0001 |
| 10th function (min) | $f = \sum_{i=1}^{N} i x_i$ | 15 | $-5 \leq x_i \leq 5$ | 0.0001 |

Table 2: Summary of results for maximizing 1$^{st}$ test function with baseline parameter values

| Population size | Mean value of best individual (Standard deviation) |
|---|---|
| 5 cells | >0.99999 (<0.000001) |
| 10 cells | 0.99976 (0.000487) |
| 20 cells | 0.86843 (0.2286) |

Table 3: Impact of reinitialization, mutation of best and Gaussian mutation for 1st test function
Mean value (Standard deviation) of best individual over 60 simulations

|  | Without reinitialization | Without mutation of the best individual | Without mutation |
|---|---|---|---|
| 5 cells | .0.99885 (0.00105) | 0.99999 (<0.000001) | 0.93164 (0.10289) |
| 10 cells | 0.97444 (0.0559) | 0.93387 (0.08567) | 0.80000 (0.18305) |
| 20 cells | 0.87492 (0.09710) | 0.74421 (0.21786) | 0.46326 (0.2856) |

These same three cases are now examined with respect to their performance when the level of mutation varies. The objective function values presented in Figure 6 represent mean values of the best individual averaged over 60 simulations. In Figure 6a, we see the effect on performance due to variation in the frequency of mutation of the best individual, when all other factors are kept constant. For the most part, algorithmic performance degrades when the period of mutation of the best individual is longer than two generations. The exception is the 5-cell configuration that appears indifferent to the frequency. Meanwhile, the 20-cell configuration is much more sensitive. On the other hand, Figure 6b shows the impact of the period of mutation (i.e. number of generations between mutations), when all other factors are kept constant. We see that the performance degrades as the mutations become more limited. This does not apply to the case with five cells, which again seems to have relatively constant performance. However, if the gap between two subsequent mutations gets even longer, the performance of the 5-cell case degrades as well.

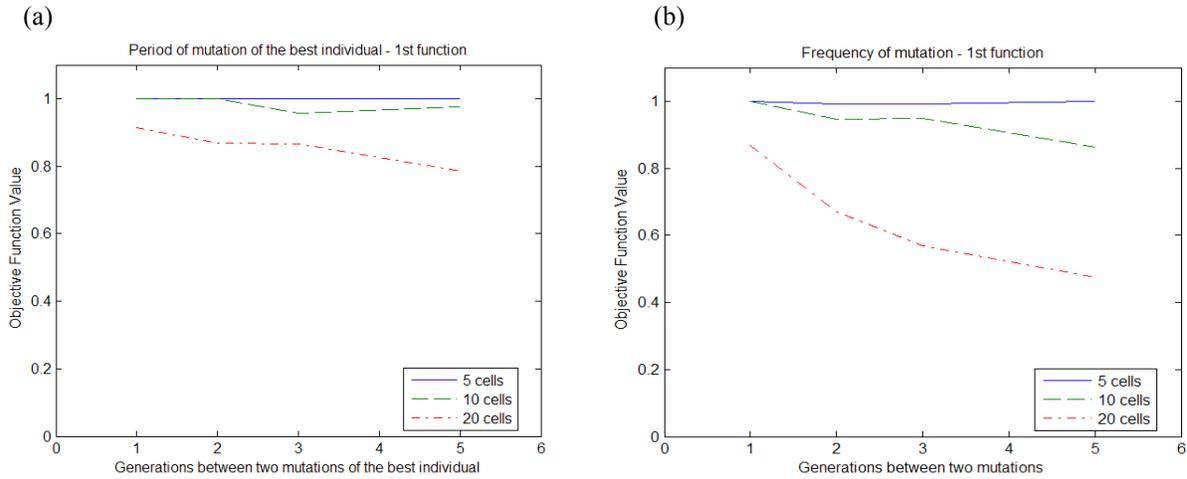

Figure 6: Period of mutation of the best individual and Gaussian mutation for 1st test function.

Figure 7 illustrates the effect of the number of mutations per generation, whenever a particular generation is selected for mutation. All other factors are kept constant at their baseline values. Notice that Figure 7 shows that an increased number of mutations gives improved results when more cells are used. This result confirms the indication from Table 2, where the 20-cell case did not perform well because of limited number of mutations.

Figure 8a shows the significance of the period of reinitialization with all other factors kept constant. We see that the optimum period is between two and five generations. Finally, Figure 8b examines the importance of the beginning point of reinitialization. The figure shows that the beginning point is not of significant importance, especially for the 5-cell case. For the other two cases, the optimum beginning for reinitialization is at approximately 25% of the total function evaluations.

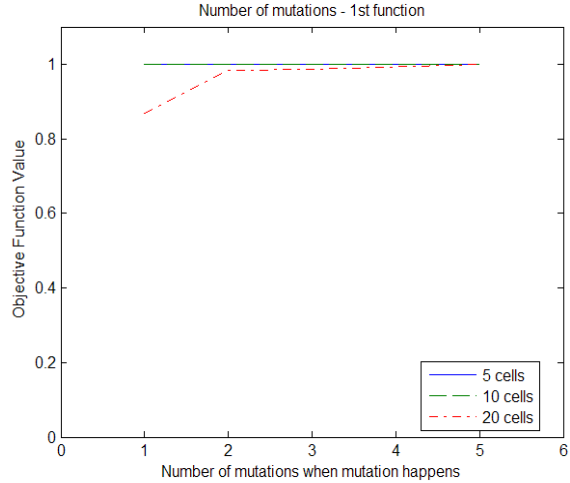

Figure 7: Impact of number of mutations per generation for 1st test function.

(a)          (b)

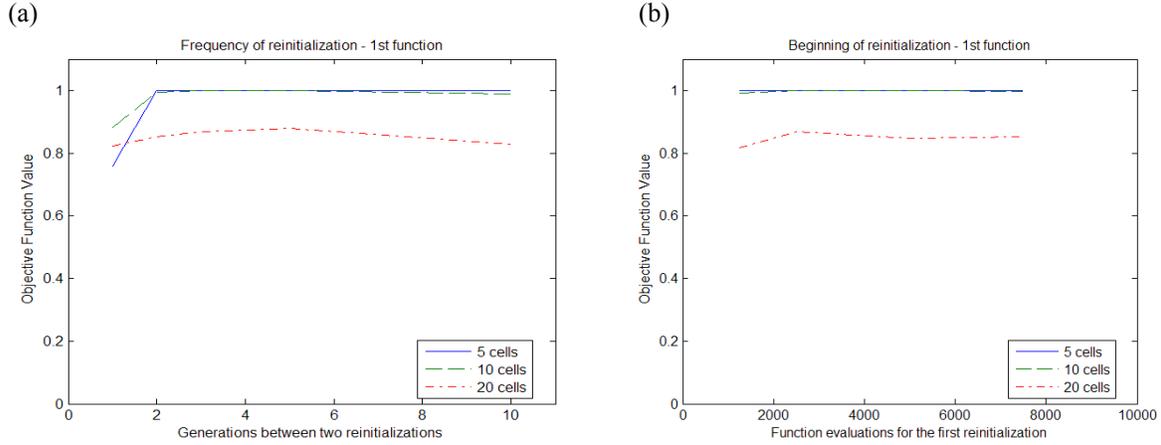

Figure 8: Impact of frequency and beginning point of reinitialization for 1st test function.

## *2nd test function*

The same procedure used in the first function will be repeated for minimization of the second test function. However, in this case, one point crossover will be employed, along with ordinary mutation. Again, five cells are considered over 2000 generations (i.e., 10000 function evaluations). Mutation of the best individual happens every 10 generations. Regular mutation, in which 15 digits are randomly changed, occurs every generation. The 15 digits correspond to five changes to each of the three substrings. Reinitialization takes place every three generations, beginning after generation 1500, which corresponds to 75% of the function evaluations. The performance with a population size of five cells then is compared with simulations having 3, 4, 10 cells and 20 cells in the population. All factors remain equal, except the beginning generation of reinitialization, which is set to 75% of the total function evaluations (e.g., at generation 750 for the 10-cell case). The mean values and standard deviations of the best individual over 60 runs are shown in Table 4 for these five cases. Once again, optimal performance was obtained for the 5-cell case. While the 10-cell case is nearly as effective, the other three cases clearly did not perform as well.

Table 4: Summary of results for minimizing 2$^{nd}$ test function with baseline parameter values

| Population size | Mean value (Standard deviation) of the best individual |
|---|---|
| 3 cells | 0.25205 (0.34014) |
| 4 cells | 0.21145 (0.27376) |
| 5 cells | 0.11405 (0.15707) |
| 10 cells | 0.12458 (0.13828) |
| 20 cells | 0.32825 (0.53592) |

Next, we consider the cases where 1) no reinitialization is invoked, 2) no mutation of the best individual is employed and 3) no ordinary mutation is included. The results shown in Table 5 represent mean values and standard deviations of the best individual averaged over 60 runs for cases with 5, 10 and 20 cells. From Table 5 we see that all the three factors have beneficial impact, except maybe from the reinitialization for the 20-cell case. The performance of the algorithm degrades substantially, especially when mutation is eliminated. Mutation of the best individual is quite important for the 5-cell case, where the performance of the algorithm is reduced more than for the cases with populations of 10 and 20 cells.

Table 5: Impact of reinitialization, mutation of best and Gaussian mutation for 2$^{nd}$ test function
Mean value (Standard deviation) of best individual over 60 simulations

|  | Without reinitialization | Without mutation of the best individual | Without mutation |
|---|---|---|---|
| 5 cells | 0.24166 (0.39004) | 0.29057 (0.28248) | 1.43486 (4.2377) |
| 10 cells | 0.21401 (0.225409) | 0.20893 (2.12238) | 2.98571 (0.16317) |
| 20 cells | 0.29495 (0.16317) | 0.35429 (5.710) | 3.07223 (0.49986) |

The three cases now are examined with respect to their performance when mutation, mutation of the best and reinitialization vary. In Figure 9a, we see the change in performance due to frequency of mutation of the best individual, when all other factors are kept constant. Clearly, the performance improves when the period is prolonged, and reaches a relatively constant level for periods around 5 to 10 generations in duration. This plateau is reached earlier for the case with 20 cells and, in that case, the performance is relatively poor. Figure 9b shows the impact of the period of mutation, when all other factors are kept constant. When the gap between two successive mutations gets longer, the performance of all three cases degrades. This is especially true for the case with 10 cells in the population.

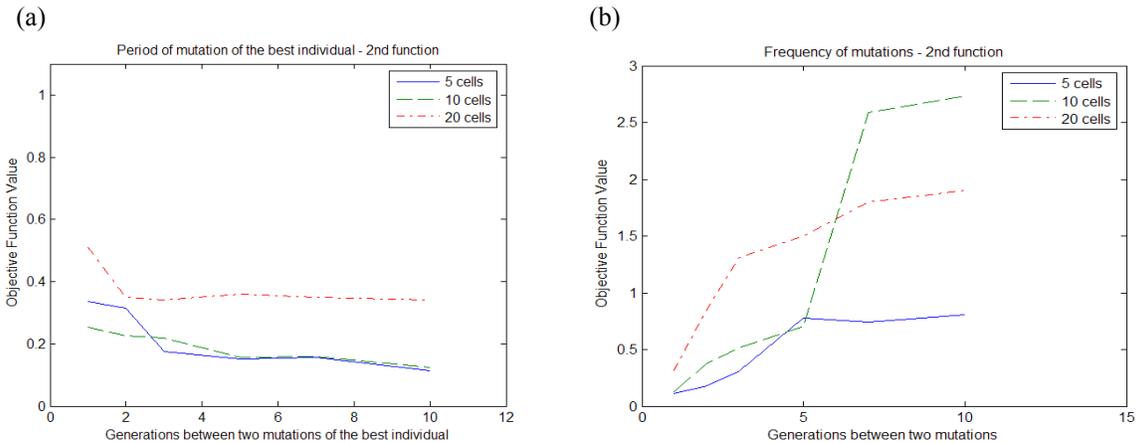

Figure 9: Period of mutation of the best individual and Gaussian mutation for 2$^{nd}$ test function.

Figure 10 shows performance as the number of mutations per generation varies, whenever the generation is chosen for mutation. All other factors are kept constant at their baseline values. We see that the optimum region is around 4 to 7 mutations per generation. Fewer mutations lead to reduced efficiency of the algorithm, while more mutations cannot be effectively managed by the cells, although in most cases the decrease in performance is not so great. For an optimal level of mutations, we see that the 20-cell case can give results that are very close to the best results of the cases with smaller population sizes.

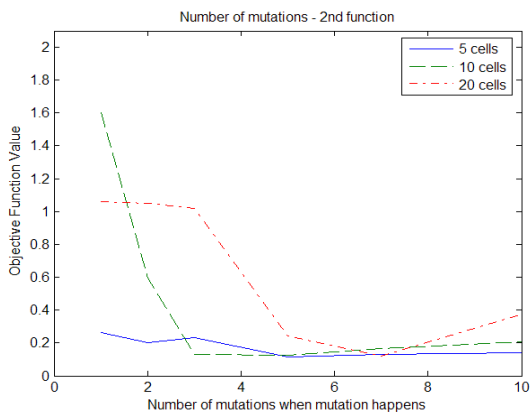

Figure 10: Impact of number of mutations per generation (whenever mutations occur) for the 2$^{nd}$ test function.

Figure 11a shows the significance of the period of reinitialization, again, with all other factors constant. We see that the optimum period is between 2 and 5 generations. Finally, Figure 11b shows the importance of the beginning point of reinitialization. The figure indicates that the beginning point can have a significant impact on results, especially if reinitialization starts too early.

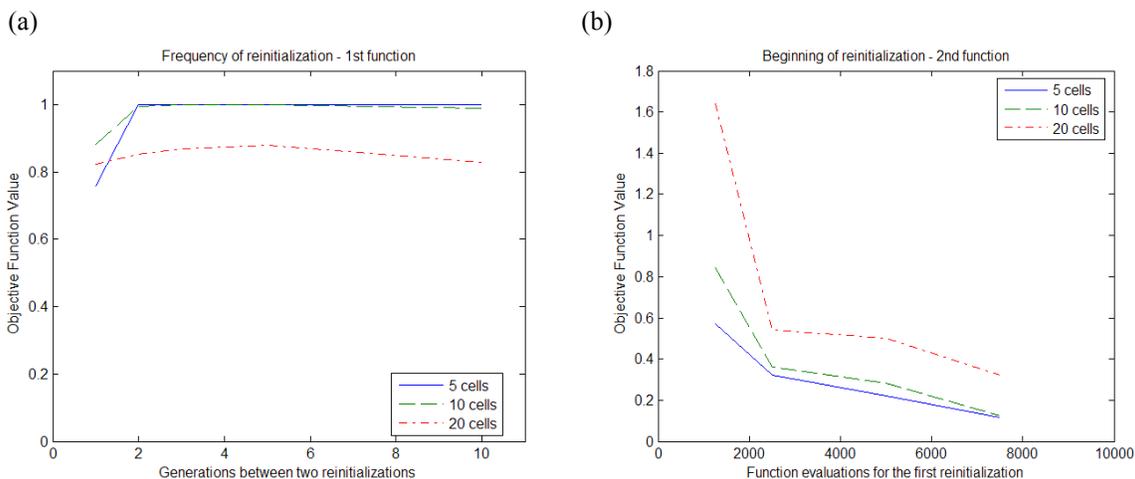

Figure 11: Impact of frequency and beginning point of reinitialization for 2$^{nd}$ test function.

*Conclusions from the study of the two first test functions*

From the previous parametric studies, it is obvious that sufficient mutation (both regular mutation and mutation of the best individual) is of critical importance for the performance of the algorithm. Reinitialization also can be important. It allows the algorithm to concentrate on the optimum-so-far area after a number of mutations that might not be very successful, enhancing in that way the exploitation abilities of the algorithm. Five cells seem to provide a robust configuration, as opposed to populations with a lesser or greater number of cells. This does not mean, however, that more cells cannot perform well. In those cases, there is just a need for an increased number of mutations per generation. On the other hand, performance degraded significantly for populations with less than five cells.

*Comparison with ordinary GA*

The remaining eight functions, as well as the baseline configuration for the first two functions, are now tested against PIKAIA (Charbonneau and Knapp, 2007), an ordinary GA. For the proposed algorithm, a population of five cells is employed for all of these tests. The functions 1 and 2 run for 10000 function evaluations, the functions 3-7 run for 20000 function evaluations, and the functions 8-10 for 5000 evaluations. Functions 3-10 use variable-to-variable crossover, mutation of the best individual every two generations and regular mutation every generation. All mutations are Gaussian and for the regular mutation case, only one mutation occurs per generation. Reinitialization for functions 3-7 occurs every three generations after 25% of function evaluations are completed (1000 generations). Meanwhile, for functions 8-10, since the total number of evaluations is small, reinitialization is triggered every third generation, but it now starts after 50% of the function evaluations are complete (i.e., after 500 generations).

For functions 1 and 2 with PIKAIA, the recommendations for these two functions found in Koumousis and Katsaras (2006) are used. On the other hand, for functions 3-10, a population of 100 individuals is utilized in PIKAIA, along with two different rates of mutation 0.005 and 0.045. Then, results from the best performing case are taken for the comparison.

The results from the comparison are shown in Table 6, where it can be seen that the proposed algorithm outperforms PIKAIA for all cases. The results presented in this table represent the mean values and standard deviations of the best individual over 60 runs. The [1] and [2] in brackets in the last column for PIKAIA correspond to 0.005 and 0.045 rates of mutation, respectively. Figures 12a-j provide further details concerning the evolution of the objective function versus generation number. All of these results are based upon a set of 60 simulations for each of the ten functions. The mean value of the best individual averaged over 60 runs is presented, along with value of the best individual found in any of those 60 runs.

Table 6: Comparison with results from PIKAIA
Mean value (Standard deviation) of best individual over 60 simulations

| | Function evaluations | Proposed algorithm | PIKAIA |
|---|---|---|---|
| 1st function (max) | 10000 | >0.99999 (<0.000001) | 0.94221 [1] |
| 2nd function (min) | 10000 | 0.11405 (0.15707) | 0.64075 [2] |
| 3rd function (min) | 20000 | 0.08607 (0.071575) | 16.14303 [1] |
| 4th function (min) | 20000 | -6284.1 (0.003882) | -6075.7 [1] |
| 5th function (min) | 20000 | 0.08869 (0.06321) | 0.28652 [1] |
| 6th function (min) | 20000 | 0.00003 (0.00008) | 0.01924 [1] |
| 7th function (min) | 20000 | 0.00002 (<0.000001) | 0.02278 [1] |
| 8th function (max) | 5000 | $9.0376 \times 10^7$ ($0.59864 \times 10^7$) | $7.9583 \times 10^7$ [2] |
| 9th function (max) | 5000 | 144.14277 (1.61992) | 130.4332 [1] |
| 10th function (min) | 5000 | 0.15672 (0.003882) | 5.98916 [1] |

(a)
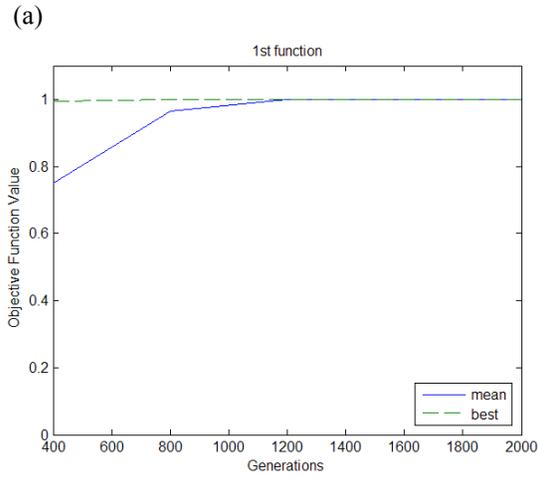

(b)
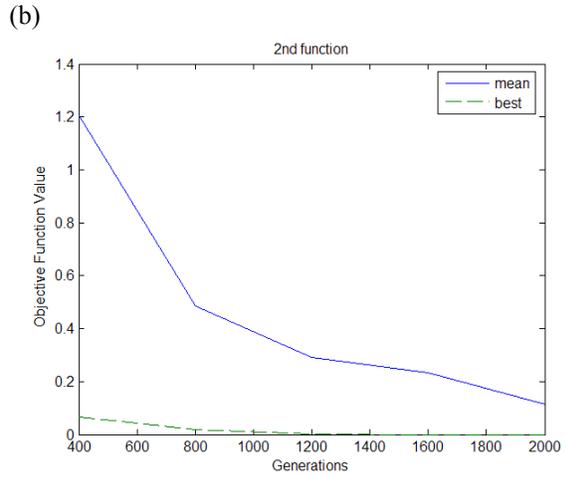

(c)
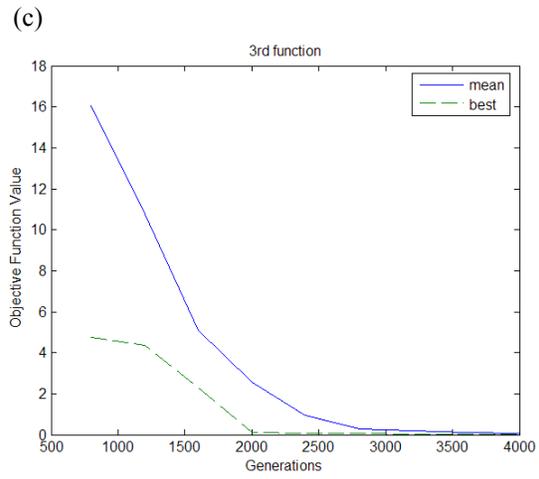

(d)
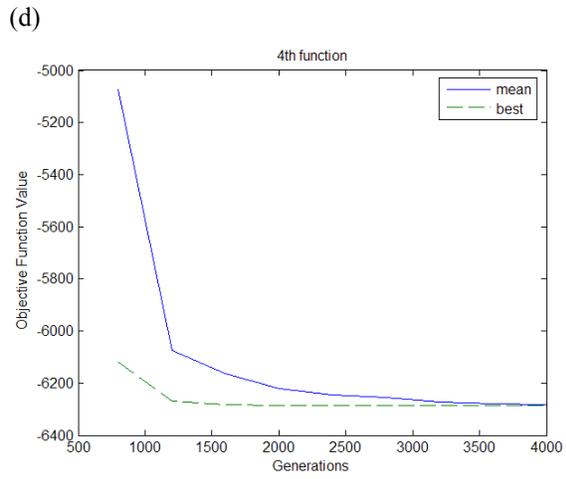

(e)
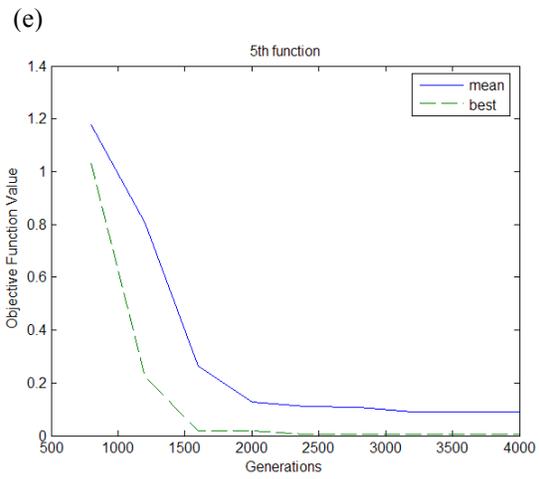

(f)
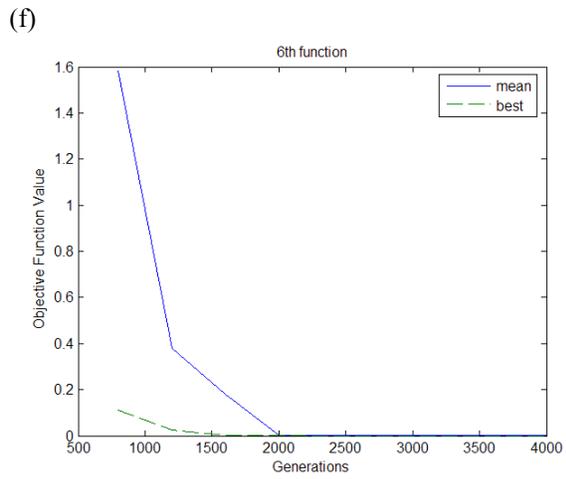

Figure 12: Evolution of the test functions versus generation number.

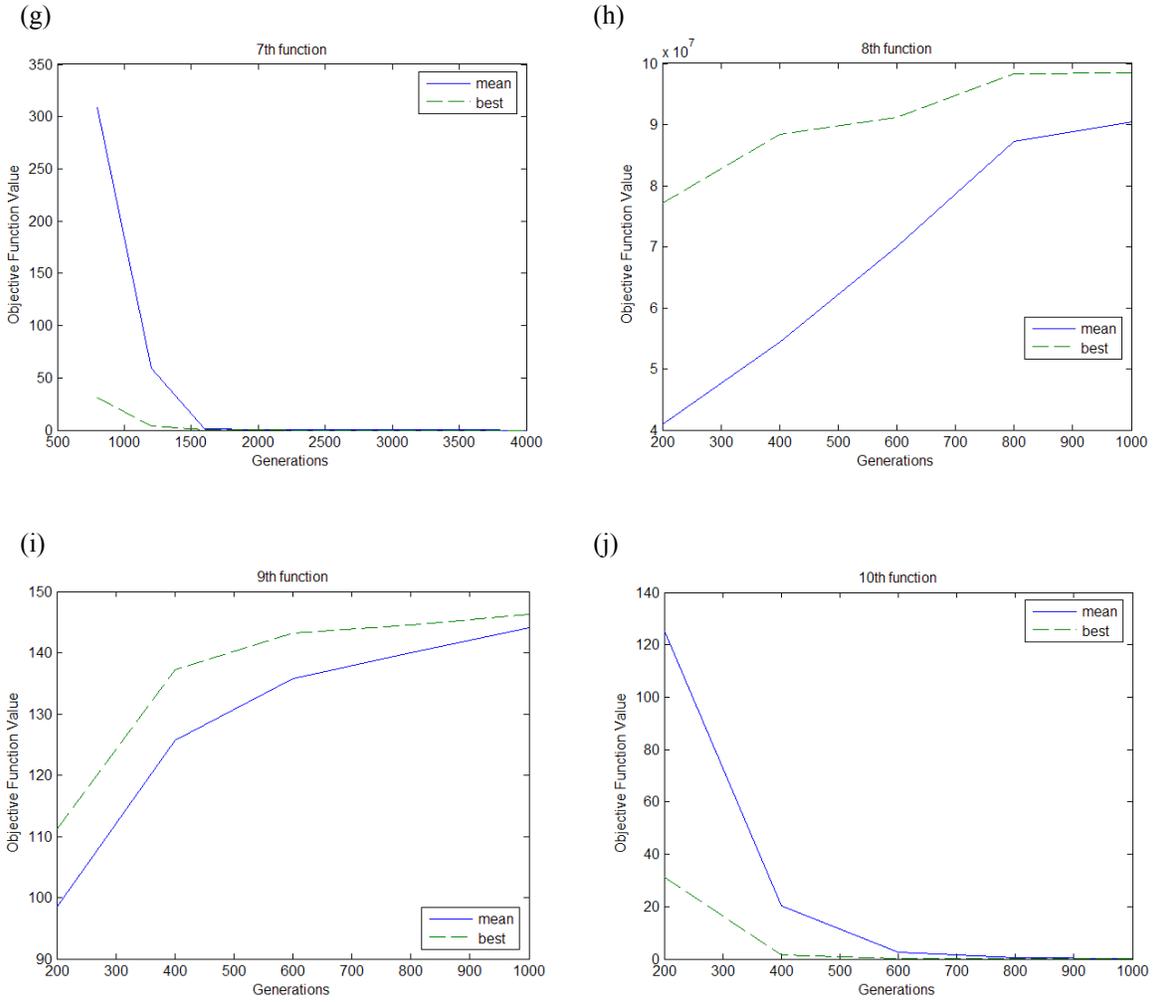

Figure 12 (continued): Evolution of the test functions versus generation number.

## 5 Structural Optimization

*Introduction*

In this section, the proposed CA-based GA approach is applied to a more practical engineering problem, involving the optimal design of structural systems. Usually the structural optimization problem is posed as the constrained minimization of the total weight of the structure. For a truss-like structure with $M$ elements and $K$ nodal degrees of freedom, this can be formulated as:

$$\min W = \sum_{m=1}^{M} \rho_m A_m L_m \tag{3}$$

with constraints:

$$d_k \leq d_{k,allowable} \quad \text{for } k = 1, 2, \ldots, K \tag{4}$$

$$\sigma_m \leq \sigma_{m,allowable} \quad \text{for } m = 1, 2, \ldots, M \tag{5}$$

where for truss member $m$, $\rho_m$ is the density of the material, $A_m$ represents the cross-sectional area and $L_m$ symbolizes the length of the member. Meanwhile, $d_k$ and $\sigma_m$ represent the corresponding displacement degree of freedom $k$ and the stress of member $m$, respectively, under a given loading condition (e.g., gravity loading, wind loading or seismic loading). The nodal displacements and element stresses are determined, in the present work, by solving a linear problem of matrix structural analysis (e.g., McGuire et al., 2000; Przemieniecki, 1985).

Before addressing this specific truss design problem, a brief review of work involving the application of GA and CA within the field of structural engineering will be provided. Goldberg and Samtani (1986) first applied GAs in structural engineering for the optimization of a benchmark ten-bar truss. More recently, Koumousis and Georgiou (1994) used a combination of genetic algorithms and logic programming to optimize a steel roof using discrete optimization. The layout of the steel truss is optimized by GAs, while the sizing optimization problem is solved by a logic program. Hajela and Lee (1995) used GAs for the topology and shape optimization of a steel truss using the concept of ground structure method and a GA with two stages. Their algorithm searched for the optimum connectivity between elements. Once the geometry of the truss is established, then the size optimization of the truss members follows. Cappello and Mancuso (2003) combined finite elements with GAs for topology and shape optimization using the homogenization theory. They used a grid-like continuum for the geometry optimization and investigated the benefit of adaptive crossover and mutation probabilities for geometry optimization. Dargush and Sant (2005) used GAs for the seismic design of buildings with passive energy dissipation devices. The optimization process included sizing and placement of passive dampers, under an uncertain seismic environment. The design was divided into eras and each era included a more complicated design than the previous one.

CA have also been used in structural analysis and structural optimization. In the field of structural engineering, Gürdal and Tatting (2000) and Tatting and Gürdal (2000) have applied CA in order to perform analysis of trusses and solids, respectively. For the truss analysis, each node of the truss became a cell of the CA lattice and the force equilibrium and Hooke's law became the rules of the CA evolution. Similarly, the continuum element is represented by a grid of equivalent truss elements. The cross sectional areas of these elements are found by using the strain energy of a truss cell and equating it with the strain energy of a continuum cell. Canyurt and Hajela (2005) proposed a GA based on CA. In that paper, they present an analysis of continuum structures by improving the treatment of the Poisson ratio as well as a CA framework for GAs, which gives better results than a traditional GA. In addition, Kicinger et al. (2004) have proposed a CA approach for the design of buildings by using morphogenetic evolutionary design.

Next, the proposed CA-based GA will be applied to the truss design problem defined in (3)-(5). Consider first the generic application of GAs to this problem. When GAs are used for structural optimization of trusses, the data string of each individual is created by the areas of the section of the structure. Therefore, if a truss has five bars, the string processed by a GA has five substrings. However, because of the inability of GAs to handle explicitly constraints, the problem is restated in the following form by introducing penalty functions:

$$\min W = \sum_{m=1}^{M} L_m \rho_m A_m \left(1 + P_d + P_s\right) \tag{6}$$

$$P_d = p_d \left( \sum_{i=1}^{K} \left\langle \frac{C_{di}}{C_{di}^{allowed}} - 1 \right\rangle \right)^2 \tag{7a}$$

$$P_s = p_s \left( \sum_{i=1}^{M} \left\langle \frac{C_{si}}{C_{si}^{allowed}} - 1 \right\rangle \right)^2 \tag{7b}$$

where $p$ represents a penalty parameter, $C$ is the value of a constraint and $\langle \cdot \rangle$ represents Macaulay brackets. Here, subscript $d$ stands for displacements and $s$ for stresses. Again, $K$ is the total number of degrees of freedom and $M$ denotes the total number of members. Equation (6) shows that the value of the objective function is the weight of the design, plus a percentage of the weight added due to violation of constraints. The penalty parameters and the magnitude of violations determine this percentage. The advantage of this formulation is that the penalty may be automatically adjusted. Many other forms of the penalty function are possible.

For the present work, the penalty function in use is a dynamic one. However, its dynamic nature has to do with limits of the constraints and not only with the penalty parameters. During the first generations, a small relaxation is imposed on the constraints. Then, after some generations, the relaxation is suddenly limited to a smaller value and the process goes on until the constraints reach the actual limits. For example, initially the limit is 2.001 instead of 2. After some iterations, it reduces to 2.0005 and then finally to 2. The concept is illustrated in Figure 13.

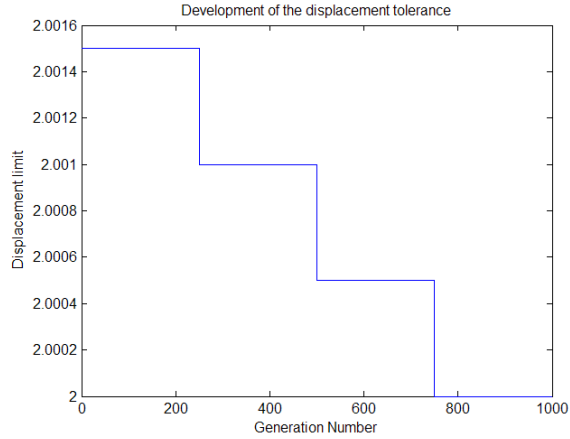

Figure 13: Evolution of displacement tolerances through the generations.

The penalty parameters $p_i$ correspond to a three segment penalty function. For the displacement constraints, these are defined as:

$$p_i = \begin{cases} P_1 & \text{if } 0 < \sum_{i=1}^{K} \left\langle \frac{C_i}{C_i^{allowed}} - 1 \right\rangle < v_1 \\ P_2 & \text{if } v_1 \leq \sum_{i=1}^{K} \left\langle \frac{C_i}{C_i^{allowed}} - 1 \right\rangle < v_2 \\ P_3 & \text{if } v_2 \leq \sum_{i=1}^{K} \left\langle \frac{C_i}{C_i^{allowed}} - 1 \right\rangle \end{cases} \tag{8}$$

If the violation is below a limit ($v_1$), the violation parameter $p_i$ assumes the value $P_1$. If the violation is greater than or equal to $v_1$, but less than $v_2$, then $p_i$ acquires the value $P_2$, with $P_2 > P_1$. If the violation is greater than or equal to $v_2$, then the penalty parameter takes the value $P_3$. Finally, $p_i$ is multiplied with the square of the sum of all violations and the penalty placed upon the design is determined. This results in a three segment penalty function, as shown in Figure 14. The third segment is used for designs that have a more than acceptable violation. All such designs are literally eliminated with this large penalty. The violation factor corresponds to $\sum_{i=1}^{K} \left\langle \dfrac{C_i}{C_i^{allowed}} - 1 \right\rangle$. The limits of the violation factor and the values of $p_i$ change after a number of generations. The violation factor becomes more restrictive and the values of $p_i$ become larger. There are two sets of parameters; one for the first generations (Figure 14a) and a second set for the last generations (Figure 14b).

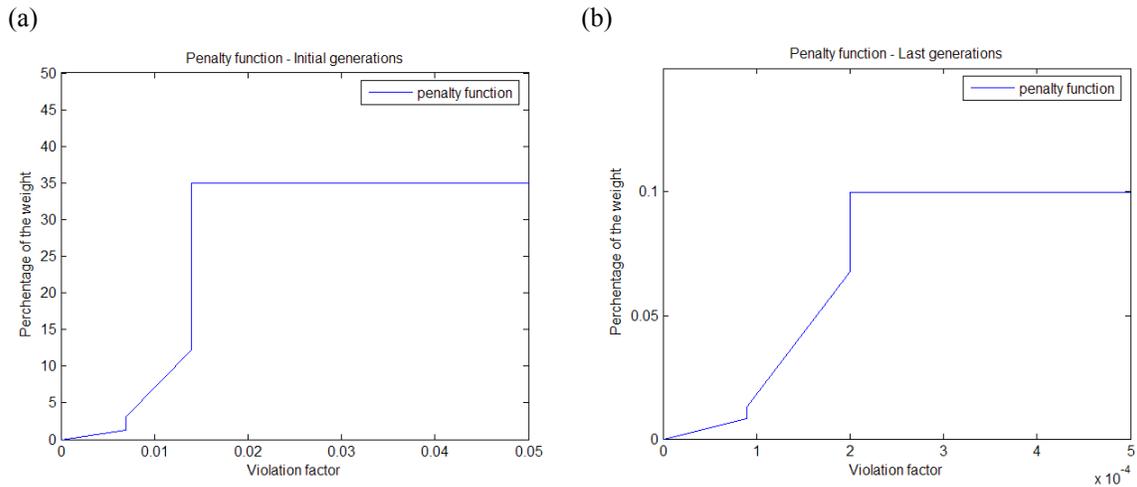

Figure 14: Evolution of penalty parameters through the generations.

*Ten bar truss*

The ten bar truss, depicted in Figure 15, is used as an example of a continuous structural optimization. Pertinent data are provided in Table 7.

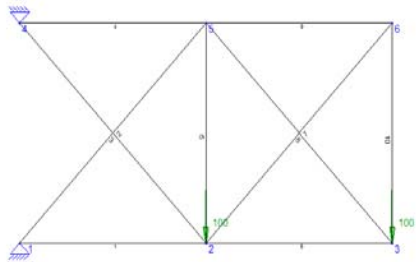

Figure 15: Ten bar truss configuration.

Table 7: Data for the ten bar truss

| | |
|---|---|
| Dimension of all vertical bars: | 914.4 cm |
| Dimension of all horizontal bars: | 914.4 cm |
| Dimension of all diagonal bars: | 1293.157cm |
| Load of all nodes: | 444.82 kN |
| Young's modulus: | 68.9476 GPa |
| Density of material: | 27.1447 kN/m$^3$ |
| Displacement constraint: | 5.08 cm |
| Stress constraint: | 172.375 MPa |
| Maximum section allowed: | 222.594 cm$^2$ |
| Minimum section allowed: | 0.6452 cm$^2$ |
| Variable step: | 0.0223 cm$^2$ |

The range of cross-sectional areas varies from 0.1 in$^2$ to 30 in$^2$ (i.e., 0.645 to 222.59 cm$^2$), while the precision of the design variables is 0.00345 in$^2$ (0.0223 cm$^2$). A population of five cells was employed again throughout and the algorithm executed 2100 generations (i.e., 10500 truss evaluations). One point crossover was used for the first third of the generations and variable to variable for the rest. One point crossover can achieve a good exploration of the design space, as shown in Hasancebi and Erbatur (2000). Mutation of the maximum individual happened every five generations, while regular mutation occurred every generation. For the first third of the generations, ordinary mutation was used, whereas for the remaining generations Gaussian mutation was invoked. Replacement of parts of some substrings by parts of older substrings occurred every ten generations. This replacement is a means to preserve good parts of older strings that are lost during the evolutionary process. This characteristic has proven to be useful, because under constrained optimization, some good parts are eliminated if the overall design violates the constraints significantly.

The best weight achieved with the proposed algorithm over 90 simulations was 5061.5 lb (22.514 kN), while the average weight obtained was 5087.75 lb (22.631 kN). Figure 16 displays the evolution of the design with minimum weight using the proposed compact CA-GA. The critical constraints for the ten bar truss problem were the vertical displacements of nodes 3 and 6 and the stress in the 5$^{th}$ bar. The vertical displacement of nodes 3 and 6 were at or just below 2 in (5.08 cm). No design with any violation of more than 10$^{-4}$ was accepted as a result. A small tolerance makes things easier for any algorithm and can reduce considerably the weight of the truss. Thus, a result with almost no tolerance is more difficult to achieve. With relatively large tolerances, the problem is changed significantly. For comparison, only works that allowed very limited constraint violations are considered in Table 8. The FEAPGEN (Camp et al., 1998) result is also based upon a GA approach to find the optimum design, while Haug and Aurora (1979) used an analytical method to calculate the optimal sections.

Table 8: Results and comparison for ten bar truss

| | Proposed algorithm | FEAPGEN | Haug and Aurora |
|---|---|---|---|
| Member 1 (cm$^2$) | 150.59 | 155.30 | 150.16 |
| Member 2 (cm$^2$) | 137.91 | 141.62 | 136.77 |
| Member 3 (cm$^2$) | 48.25 | 49.62 | 48.18 |
| Member 4 (cm$^2$) | 194.64 | 186.59 | 193.76 |
| Member 5 (cm$^2$) | 0.65 | 0.65 | 0.65 |
| Member 6 (cm$^2$) | 98.75 | 90.07 | 98.63 |
| Member 7 (cm$^2$) | 0.65 | 0.65 | 0.65 |
| Member 8 (cm$^2$) | 137.24 | 142.52 | 139.48 |
| Member 9 (cm$^2$) | 0.65 | 0.65 | 0.65 |
| Member 10 (cm$^2$) | 3.54 | 3.61 | 3.59 |
| Weight (kN) | 22.51 | 22.58 | 22.52 |

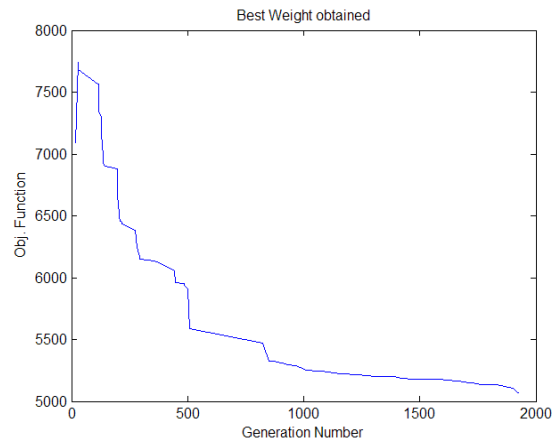

Figure 16: Minimum weight at each generation for ten bar truss.

*Seventeen bar truss*

The proposed algorithm is also tested on the continuous optimization of a seventeen bar truss. This truss was studied by Khot and Berke (1984) and is displayed in Figure 17, while characteristic data are provided in Table 9.

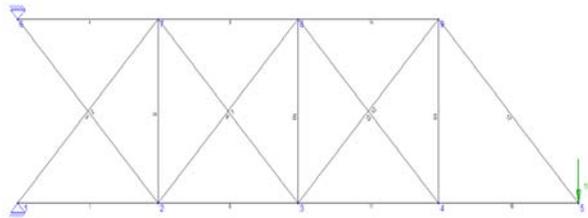

Figure 17: Seventeen bar truss configuration.

Table 9: Data for seventeen bar truss

| | |
|---|---|
| Dimension of all vertical bars: | 254 cm |
| Dimension of all horizontal bars: | 254 cm |
| Dimension of all diagonal bars: | 359.2068 cm |
| Load of all nodes: | 444.8 kN |
| Young's modulus: | 68.95 GPa |
| Density of material: | 72.735 kN/m$^3$ |
| Displacement constraint: | 5.08 cm |
| Stress constraint: | 344.75 kN |
| Maximum section allowed: | 838.76 cm$^2$ |
| Minimum section allowed: | 0.254 cm$^2$ |
| Variable step: | 0.83876 cm$^2$ |

The range of cross-sectional areas varies from 0.1 in$^2$ to 130 in$^2$ (0.254 to 838.76 cm$^2$) with a precision of 0.13 in$^2$ (0.83876 cm$^2$). The algorithm executed 2500 generations with a population of five cells (i.e., 12500 truss evaluations). Otherwise, the simulation parameters are identical to those used for the ten bar truss problem.

The best weight obtained over 30 runs was 2570 lb (11.432 kN). The weight obtained by Khot and Berke (1984) was 2580 lb, or 11.476kN (Table 10), while Adeli and Kumar (1995) reported a weight of 2593.3 lb (11.536 kN). The critical constraint for this problem was the vertical displacement of node 5. Here its value was 2.0024 in, while the same displacement in the Khot and Berke (1984) design was 2.0023 in. Table 10 provides a comparison of member sizes obtained by the proposed algorithm with those presented by Khot and Berke (1984). In this example, the displacement violation was allowed some tolerance for the sake of comparison with published results. In Figure 18, the evolution of the best weight is presented versus generation number.

Table 10: Results and comparison for seventeen bar truss

|  | Proposed algorithm | Khot & Berke [21] |
|---|---|---|
| Member 1 (cm$^2$) | 67.32 | 77.81 |
| Member 2 (cm$^2$) | 36.49 | 35.87 |
| Member 3 (cm$^2$) | 0.65 | 0.65 |
| Member 4 (cm$^2$) | 91.48 | 102.78 |
| Member 5 (cm$^2$) | 0.67 | 0.65 |
| Member 6 (cm$^2$) | 83.15 | 76.97 |
| Member 7 (cm$^2$) | 0.65 | 0.65 |
| Member 8 (cm$^2$) | 32.33 | 35.87 |
| Member 9 (cm$^2$) | 50.65 | 52.00 |
| Member 10 (cm$^2$) | 0.67 | 0.65 |
| Member 11 (cm$^2$) | 24.83 | 26.13 |
| Member 12 (cm$^2$) | 41.49 | 36.00 |
| Member 13 (cm$^2$) | 0.65 | 0.65 |
| Member 14 (cm$^2$) | 54.82 | 51.23 |
| Member 15 (cm$^2$) | 0.65 | 0.65 |
| Member 16 (cm$^2$) | 28.16 | 25.81 |
| Member 17 (cm$^2$) | 40.66 | 36.45 |
| Weight (kN) | 11.43 | 11.48 |

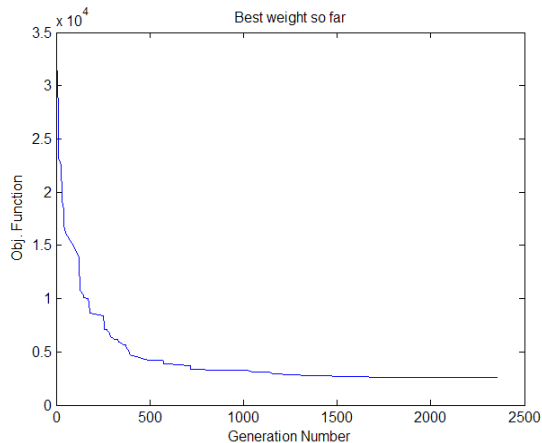

Figure 18: Minimum weight at each generation for seventeen bar truss.

# 6  Conclusion

A new localized paradigm for GAs is developed. The proposed algorithm makes use primarily of the self-organizing capabilities of CA. An individual of the population exchanges information only with its immediate neighbors and no global statistics of the whole population are used to evolve the system. The proposed algorithm uses very high rates of mutation and this mutation is crucial for successful optimization. Ten test functions were presented and two benchmark trusses were optimized. The optimization of all functions was robust with the optimum value reached in a limited number of evaluations. For both benchmark trusses, the proposed algorithm achieved lightweight designs comparable or better than those published in the literature.

The range of problems considered here suggests that the proposed algorithm has potential as a robust optimization tool. Applications to other engineering design problems are underway, including situations involving multi-objective optimization and noisy fitness functions.

**References**

Adeli, H. and Kumar, S. (1995). Distributed genetic algorithm for structural optimization. *J. Aerospace Engineering*, ASCE, 8: 156-163.

Alba , E., Giacobini M., Tomassini M. and Romero S. (2002). Comparing synchronous and asynchronous cellular genetic algorithms. In *Proceedings of 7$^{th}$ International Conference on Parallel Problem Solving from Nature*, 601-610, Springer Verlag.

Alba, E. and Luque, G. (2004). Growth curves and takeover time in distributed evolutionary algorithms. In *Proceedings of GECCO 2004*, 864-876, Springer Verlag.

Alba, E. and Tomassini, M. (2002). Parallelism and evolutionary algorithms. *IEEE Trans. Evol. Comp.*, 6: 443-462.

Alba, E. and Troya, J.M. (2000). Cellular evolutionary algorithms: Evaluating the influence of ratio. In *Proceedings of 6$^{th}$ International Conference on Parallel Problem Solving from Nature*, 29-38, Springer Verlag.

Camp, C., Pezeshk, S. and Cao, G. (1998). Optimized design of two-dimensional structures using a genetic algorithm. *J. Struct. Eng.*, ASCE 124: 551-559.

Canyurt, O.E. and Hajela, P. (2005). A Cellular framework for structural analysis and optimization. *Comput. Methods Appl. Mech. Engrg.*, 194: 3516–3534.

Cappello, F. and Mancuso, A. (2003). A genetic algorithm for combined topology and shape optimization. *Computer-Aided Design*, 35: 761-769.

Charbonneau, P. and Knapp, B. (2007). PIKAIA, [Available online at: http://www.ecy.wa.gov/programs/eap/models.html].

Collins, R. and Jefferson, D. (1991). Selection in massively parallel genetic algorithms. In *Proc. 4$^{th}$ Int. Conf. on Genetic Algorithms*, L. Booker and R. Belew, Eds., 249-256, Morgan Kaufmann.

Cui, Z.-H., Zeng, J.-C. and Xu, Y.-B. (2003). Master-to-slave nonlinear genetic algorithm. In *Proceedings of 42$^{nd}$ IEEE Conference on  Decision and Control*, IEEE Press.

Davidor, Y. (1991). A naturally occurring niche and species phenomenon: The model and first results. In *Proc. 4$^{th}$ Int. Conf. on Genetic Algorithms*, 257-263, Morgan Kaufmann.

Dargush, G. F. and Sant, R. S. (2005). Evolutionary aseismic design and retrofit of structures with passive energy dissipation. *Earthquake Engng. Struct. Dyn.*, 34: 1601-1626.

Eiben, A. E., Aarts, E. H. and van Hee, K. M. (1991). Global convergence of genetic algorithms: A Markov chain analysis. In *Parallel Problem Solving from Nature*, Schwefel, H. P. and Manner, R., Eds., 4-12, Springer-Verlag.

Giacobini, M., Tomassini, M., Tettamanzi, A. G. B. and Alba, E. (2005). Selection intensity in cellular evolutionary algorithms for regular lattices. *IEEE Trans. Evol. Comput.*, 9: 489-505.

Goldberg, D.E. (1989). *Genetic algorithms in search, optimization, and machine learning*, Addison-Wesley.

Goldberg, D.E. and Samtani, M.P. (1986). Engineering optimization via genetic algorithm. In *Proceedings of ASCE Conference in Electronic Computation*, 471-482.

Gorges-Schleuter, M. (1989). ASPARAGOS an asynchronous parallel genetic optimisation strategy, In *Proc. 3$^{rd}$ Int. Conf. on Genetic Algorithms*, J. D. Schaffer, Ed., 422-427, Morgan Kaufmann.

Guo, G. and Yu, S. (2003). Evolutionary parallel local search for function optimization. *IEEE Trans. Syst., Man, Cybern. B*, 33: 864-876.

Gűrdal, Z. and Tatting, B.T. (2000). Cellular automata for truss structures with linear and nonlinear response. In *Proceedings of 41st AIAA/ ASME/ASCE/AHS/ASC Conf. Structures, Structural Dynamics, and Materials,* AIAA Paper 2000-1580.

Hajela, P. and Lee, E. (1995). Genetic algorithms in truss topological optimization. *Int. J. Solids Structures*, 32: 3341-3357.

Hasancebi, O. and Erbatur, F. (2000). Evaluation of crossover techniques in genetic algorithm based optimum structural design. *Comput. Struct.*, 78: 435–448.

Haug, E.J. and Aurora, J.S. (1979). *Applied optimal design*, John Wiley & Sons.

Hillis, D. (1990). Co-evolving parasites improve simulated evolution as an optimization procedure. *Physica D*, 42: 228-234.

Holland, J. H. (1975). *Adaptation in natural and artificial systems*, University of Michigan Press.

Ishibuchi, H. and Murata, T. (1998). A multi-objective genetic local search algorithm and its applications to flowshop scheduling. *IEEE Trans. Syst., Man, Cybern. C*, 28: 392-403.

Kawamura, S., Yoshida, T., Minamoto H. and Hossain, Z. (2006). Simulation of the nonlinear vibration of a string using the cellular automata based on the reflection rule. *Applied Acoustics*, 67: 93–105.

Kicinger, R., Arciszewski, T. and DeJong, K. (2004). Morphogenesis and structural design: Cellular automata representations of steel structures in tall buildings. In *Proceedings of the Congress on Evolutionary Computation*.

Khot, N.S. and Berke, L. (1984). Structural optimization using optimality criteria method, New directions in optimum structural design. In *New Directions in Optimum Structural Design*, Atrek, E., Gallagher, R.H., Ragsdell, K.M., and Zienkiewicz, O.C., Eds., John Wiley.

Koumousis, V.K. and Georgiou, P.G. (1994). Genetic algorithms in discrete optimization of steel truss roofs, *J. Comp Civil Eng.,* 8: 309–328.

Koumousis, V.K. and Katsaras, C.P. (2006). A saw-tooth genetic algorithm combining the effects of variable population size and reinitialization to enhance performance. *IEEE Trans. Evol. Comput.*, 10: 19-28.

Krishnakumar, K. (1989). Micro-genetic algorithms for stationary and non-stationary function optimization. *Intell. Contr. Adapt. Syst.*, 1196: 289-96.

Leung, Y., Gao, Y. and Xu, Z.–B. (1997). Degree of population diversity - A perspective on premature convergence in genetic algorithms and its Markov chain analysis, *IEEE Trans. Evol. Comp.*, 8: 1165-1176.

Manderick, B. and Spiessens, P. (1989). Fine-grained parallel genetic algorithms. In *Proc. 3$^{rd}$ Int. Conf. on Genetic Algorithms*, J.D. Schaffer, Ed., Morgan Kaufmann.

McGuire, W., Gallagher, R.H. and Ziemian, R.D. (2000). *Matrix structural analysis*, John Wiley & Sons.

Mitchell, M. (1996). *An introduction to genetic algorithms*, MIT Press.

Mühlenbein, H. (1991). Evolution in time and space: The parallel genetic algorithm. In *Foundations of Genetic Algorithms*, G. Rawlins, Ed., 316-337, Morgan Kaufmann.

von Neumann, J. (1966). *The theory of self-reproducing automata*, University of Illinois Press.

Prügel-Bennett, A. and Shapiro, J. L. (1994). Analysis of genetic algorithms using statistical mechanics. *Phys. Rev. Letters*, 72:1305-1309.

Przemieniecki, J.S. (1985). *Theory of matrix structural analysis*, Dover.

Rudolph, G. (2000). On takeover times in spatially structured populations: Array and ring. In *Proceedings of the 2nd Asia-Pacific Conference on Genetic Algorithms and Applications*, Lai, K. K., Katai, O., Gen, M., and Lin, B., Eds., 144–151, Global-Link Publishing Company.

Sarma, J. and DeJong, K. (1996). An analysis of the effects of neighborhood size and shape in local selection algorithms. *Lecture Notes in Computer Science*, 1141, 236-244.

Schönfisch, B. and de Roos, A. (1999). Synchronous and asynchronous updating in cellular automata. *BioSystems*, 51, 123-143.

Sipper, M. Tomassini, M. and Capcarrere. M.S. (1998). Evolving asynchronous and scalable non-uniform cellular automata. In *Proc. Int. Conf. on Artificial Neural Networks and Genetic Algorithms* (ICANNGA97), 67-71, Springer-Verlag, Vienna.

Suzudo, T. (2004). Searching for pattern-forming asynchronous cellular automata – An evolutionary approach. *Lecture Notes in Computer Science*, 3305, 151-160.

Tatting, B. and Gűrdal, Z. (2000). Cellular automata for design of two-dimensional continuum structures. In *8th AIAA/ USAF/NASA/ISSMO Symposium on Multidisciplinary Analysis and Optimization*, AIAA Paper 2000-4832.

Tomassini, M. (1993). The parallel genetic cellular automata: Application to global function optimization, In *Proc. Int. Conf. Artif. Neural Netw. Genetic Algorithms*, R. Albrecht, C. Reeves, and N. Steele, Eds, 385–391.

Ulam, S. (1952). Random processes and transformations. In *Proceedings of International Congress of Mathematicians*, 2: 85-87.

Venkatraman, S. and Yen, G.G. (2005). A generic framework for constrained optimization using genetic algorithms. *IEEE Trans. Evol. Comp.*, 9: 424-435.

Weaver, W. (1948). Science and complexity. *American Scientist*, 36: 536.

Whitley, D. (1993). Cellular genetic algorithms, In *Proc. 5th Int. Conf. Genetic Algorithms*, S. Forrest, Ed., 658.

Wiener, N. and Rosenbluth, A. (1946). The mathematical formulation of the problem of conduction of impulses in a network of connected excitable elements, specifically in cardiac muscle. *Arch. Inst. Cardiol. Mex.*, 16: 205-265.

Wolfram, S. (1994). *Cellular automata and complexity: Collected papers*, Addison-Wesley.

Wolfram, S. (2002). *A new kind of science*, Wolfram Media.